\documentclass[letterpaper]{article} 
\usepackage{aaai2026}  
\usepackage{times}  
\usepackage{helvet}  
\usepackage{courier}  
\usepackage[hyphens]{url}  
\usepackage{graphicx} 
\usepackage{natbib}  
\usepackage{caption} 
\usepackage{algorithm}
\usepackage{algorithmic}

\usepackage{amsmath}
\usepackage{subcaption}
\usepackage{multirow}
\usepackage{booktabs}
\usepackage{graphicx}
\usepackage{subfig}
\usepackage{newfloat}
\usepackage{listings}
\usepackage{graphicx}
\usepackage{amsfonts}
\usepackage{epstopdf}
\usepackage{newfloat}
\usepackage{listings}
\DeclareCaptionStyle{ruled}{labelfont=normalfont,labelsep=colon,strut=off} 
\floatstyle{ruled}
\newfloat{listing}{tb}{lst}{}
\floatname{listing}{Listing}
\nocopyright
\title{RLMR: Reinforcement Learning with Mixed Rewards for Creative Writing}
\author{
    Jianxing Liao\textsuperscript{\rm 1}\thanks{Work done when these authors interned at Tencent.},
    Tian Zhang\textsuperscript{\rm 1},
    Xiao Feng\textsuperscript{\rm 1}\thanks{Corresponding author.},
    Yusong Zhang\textsuperscript{\rm 1},
    Rui Yang\textsuperscript{\rm 1},\\
    Haorui Wang\textsuperscript{\rm 1}\footnotemark[1],
    Bosi Wen\textsuperscript{\rm 2}\footnotemark[1],
    Ziying Wang\textsuperscript{\rm 3}\footnotemark[1],
    Runzhi Shi\textsuperscript{\rm 3}\footnotemark[1]
}
\affiliations{
    \textsuperscript{\rm 1}Tencent Hunyuan Team  
    \textsuperscript{\rm 2}Tsinghua University 
    \textsuperscript{\rm 3}Peking University\\
    jianxliao@tencent.com, alicexfeng@tencent.com\footnotemark[2]
}

\usepackage{bibentry}

\begin{document}

\maketitle

\begin{abstract}
Large language models are extensively utilized in creative writing applications. Creative writing requires a balance between subjective writing quality (e.g., literariness and emotional expression) and objective constraint following (e.g., format requirements and word limits). Existing methods find it difficult to balance these two aspects: single reward strategies fail to improve both abilities simultaneously, while fixed-weight mixed-reward methods lack the ability to adapt to different writing scenarios. To address this problem, we propose Reinforcement Learning with Mixed Rewards (RLMR), utilizing a dynamically mixed reward system from a writing reward model evaluating subjective writing quality and a constraint verification model assessing objective constraint following. The constraint following reward weight is adjusted dynamically according to the writing quality within sampled groups, ensuring that samples violating constraints get negative advantage in GRPO and thus penalized during training, which is the key innovation of this proposed method. We conduct automated and manual evaluations across diverse model families from 8B to 72B parameters. Additionally, we construct a real-world writing benchmark named WriteEval for comprehensive evaluation. Results illustrate that our method achieves consistent improvements in both instruction following (IFEval from 83.36\% to 86.65\%) and writing quality (72.75\% win rate in manual expert pairwise evaluations on WriteEval). To the best of our knowledge, RLMR is the first work to combine subjective preferences with objective verification in online RL training, providing an effective solution for multi-dimensional creative writing optimization.
\end{abstract}


\section{Introduction}

Large language models (LLMs) are widely applied to creative writing tasks, from traditional poetry composition to modern fiction generation, and from literary scriptwriting to commercial copywriting, fulfilling diverse writing demands across domains and genres. To further enhance LLM performance in creative writing tasks, reinforcement learning techniques have been widely applied during the post-training phase. Through methods such as Group Relative Policy Optimization (GRPO), researchers aim to guide models toward generating higher-quality creative content through reward signals.

However, existing reinforcement learning reward strategies suffer from fundamental limitations. The evaluation criteria for creative writing are inherently dual in nature: on one hand, they require assessing subjective writing qualities such as literariness, emotional expression, and originality; on the other hand, they necessitate verifying objective constraint following, including length constraints, format requirements, and specific writing styles. Different creative writing scenarios exhibit significant variations in their emphasis on subjective versus objective evaluation.

\begin{figure*}[t]
\centering
\includegraphics[width=1.8\columnwidth]{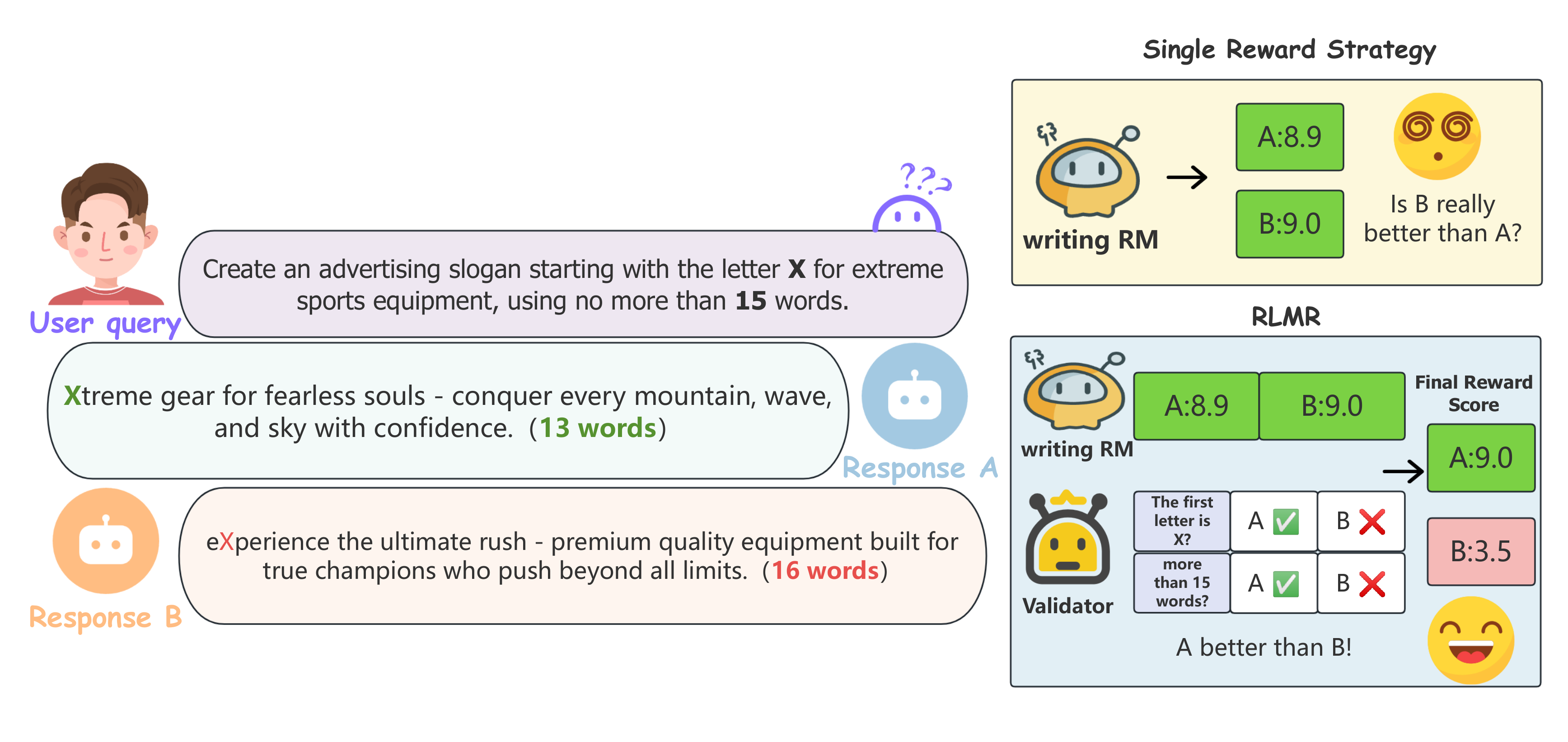} 
\caption{Comparison of single reward strategy versus our mixed RLMR approach. Given a task requiring an advertising slogan starting with "X" using no more than 15 words, Response A follows constraints but scores lower (8.9), while Response B violates constraints but scores higher (9.0). Single reward strategies incorrectly prefer Response B, while our RLMR combines writing quality and instruction following signals to correctly identify Response A as superior through dynamic penalty adjustments.}
\label{fig:introduction}
\end{figure*}

Current reward strategies face two major challenges. First, single reward strategies struggle to simultaneously optimize both subjective and objective dimensions. As illustrated in Figure~\ref{fig:introduction}, under single-signal strategies, reward models only score writing quality without reflecting constraint following. Second, existing multi-reward signal fusion strategies typically employ fixed-weight summation. Such fixed-weight mechanisms fail to dynamically adjust weights based on actual sample performance within groups, making them unsuitable for different writing scenarios.

To address these issues, we propose Reinforcement Learning with Mixed Rewards (RLMR), a dynamic mixed-reward framework for creative writing. By coupling a writing reward model for evaluating subjective writing quality with a constraint verification model for assessing objective constraint following, we implement an adaptive mechanism that dynamically allocates reward weights based on constraint following within sampled group responses. Unlike existing methods that use fixed-weight fusion, our core innovation lies in dynamically adjusting the constraint following reward weight according to writing quality within sampled groups. This ensures that samples violating constraints receive negative advantage values in GRPO calculations, thereby being systematically penalized during policy gradient updates.

To validate our method's effectiveness, we conducted training on various scales of Qwen and DeepSeek model families and performed both automated and manual evaluations on multiple creative writing and instruction-following benchmarks. RLMR shows substantial gains in both writing quality and constraint following compared to single-reward and linear weighting baseline methods. Manual evaluation confirms significant preference for our approach over traditional strategies. These results effectively validate that our method resolves the trade-off between subjective and objective evaluation criteria in creative writing optimization.

Our key contributions include:
\begin{enumerate}
\item Identifying the inherent limitations of single reward signals and fixed-weight mixing strategies in creative writing tasks.
\item Proposing RLMR and developing a dynamic reward adjustment mechanism that ensures constraint-violating samples receive negative advantages during training, enabling better balance between writing quality and constraint following among multiple reward signals.
\item Demonstrating consistent improvements across diverse model families and scales through comprehensive automated and manual evaluations, proving the effectiveness of our method.
\end{enumerate}
\begin{figure*}[h!]
\centering
\includegraphics[width=2\columnwidth]{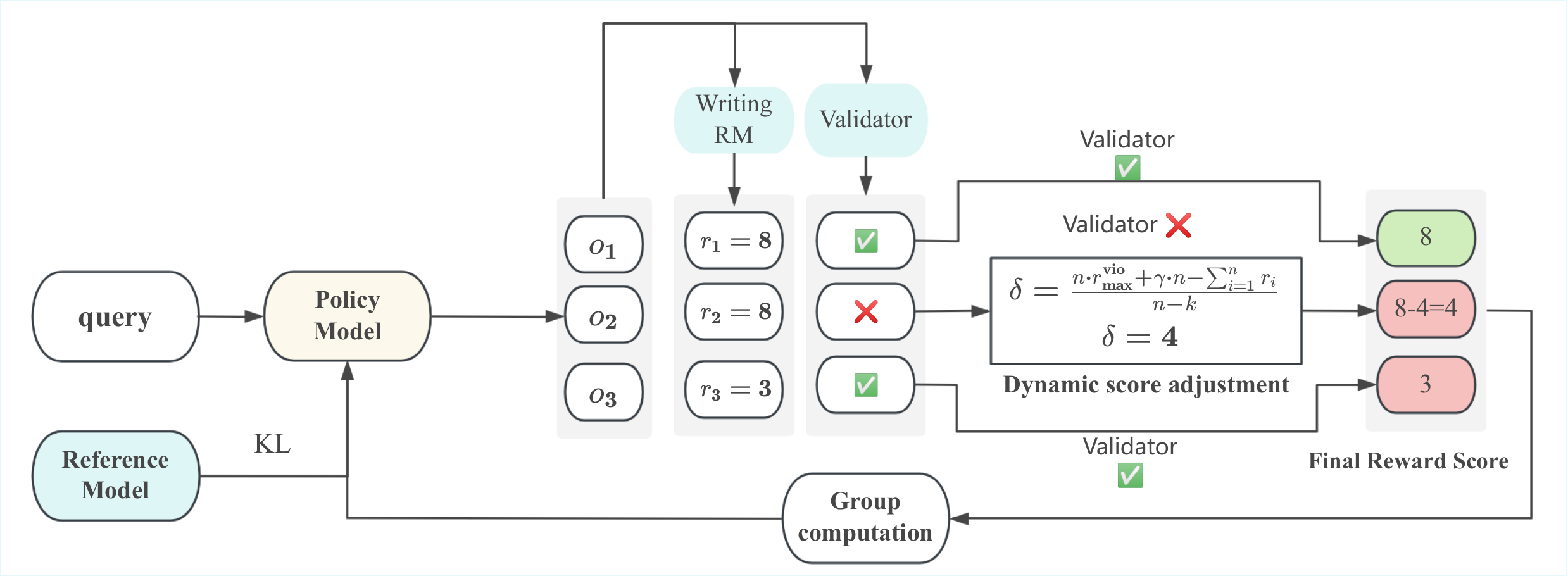} 
\caption{Overview of our Dynamic Mixed-Reward GRPO Framework. The policy model generates responses ($o_1, o_2, o_3$) evaluated by both writing quality (Writing RM) and constraint compliance (Validator). In this example: $n=3$ (total samples), $r_{\max}^{\text{vio}}=8$ (highest reward among violating samples), $\gamma=1$ (minimum gap below the mean), $k=1$ (number of violating samples), $\sum_{i=1}^{n} r_i =19$ (sum of original rewards). The framework calculates penalty $\delta = 4$ and deducts it from violating samples ($o_2: 8 \rightarrow 4$). After adjustment around mean=5, only high-quality compliant samples ($o_1$) receive positive gradients (green), while both low-quality samples ($o_3$) and constraint-violating samples ($o_2$) receive negative gradients (red).}
\label{fig:framework}
\end{figure*}
\subsection{Related Work}

To further improve LLM performance and align it with human preferences, reinforcement learning, especially RLHF, has become a mainstream optimization approach. Algorithms such as Proximal Policy Optimization (PPO) ~\cite{ouyang2022training} and Group Relative Policy Optimization (GRPO)~\cite{shao2024deepseekmath} are widely used to align LLM behavior with human preferences. PPO ensures training stability by limiting the extent of policy updates through clipped probability ratios, but requires separate value function training which increases computational overhead. GRPO optimizes policy gradients by estimating baselines from sampled groups, avoiding the need for separate value function training while maintaining competitive performance. Given GRPO's computational efficiency and effectiveness in creative writing scenarios, we choose it as our reinforcement learning framework.

Mixed reward strategies have become increasingly important in reinforcement learning, integrating multi-dimensional reward signals to guide model training more comprehensively. Peng et al.~\cite{peng2025agentic} proposed the Agentic Reward Modeling framework, which combines human preference rewards with verifiable correctness signals (factuality and instruction following) to provide more reliable rewards for large language models. Jia et al.~\cite{jia2025writingzero} introduced Writing-Zero, proposing a writing-principle-based pairwise Generative Reward Model (GRM) that leverages self-principled critique to transform subjective assessments into reliable, verifiable rewards for creative writing tasks. Wu et al.~\cite{wu2025longwriter} developed LongWriter-Zero framework for ultra-long text generation, employing specialized reward models targeting length control, writing quality, and structural formatting with a composite reward function that averages individual advantages to balance multiple reward dimensions. 

However, these existing mixed reward approaches all rely on fixed-weight fusion mechanisms, which suffer from fundamental limitations. First, fixed weights cannot adapt to varying constraint compliance patterns within different sample groups. When most responses in a group violate constraints, fixed-weight strategies still assign positive gradients to high-quality but constraint-violating samples, contradicting creative writing requirements. Second, the relative importance between subjective quality assessment and objective constraint following cannot be accurately determined, making weight assignment difficult. To address these issues, we propose a dynamic mixed-reward GRPO framework that adaptively adjusts penalty weights based on actual constraint compliance performance within each sampled group, ensuring constraint-violating samples consistently receive negative advantages during training. This dynamic adjustment approach is better suited for creative writing tasks.

\section{RLMR Framework for Creative Writing}

To effectively combine subjective and objective reward signals, we propose a mixed-reward GRPO framework. This framework integrates a writing reward model for evaluating writing quality with a verification model for assessing instruction compliance. By adjusting reward scores based on verification results, we achieve improved instruction-following capability while maintaining writing quality.

\subsection{Reward Models}
Our RLMR framework employs two reward models: a writing reward model that evaluates subjective writing quality and a constraint verification model that assesses objective compliance with task requirements. 
\paragraph{Writing Reward Model.}
The writing reward model $r_{\text{write}}$ evaluates the overall quality of creative writing outputs. We train this model on a large language model using human-annotated preference pairs $(y_w, y_l)$ for creative writing prompts $x$. Following the Bradley-Terry preference model, we optimize:

\begin{equation}
\mathcal{L}_{\text{write}} = -\mathbb{E}_{(x,y_w,y_l) \sim \mathcal{D}} \left[ \log \sigma(r_{\text{write}}(x, y_w) - r_{\text{write}}(x, y_l)) \right]
\end{equation}

where $y_w$ and $y_l$ denote preferred and non-preferred responses, and $\sigma$ is the sigmoid function. Unlike general reward models, our writing reward model captures creative writing features including literary expression, emotional depth, originality, narrative coherence, and stylistic maturity.

\paragraph{Constraint Verification Model}
The verification model identifies constraint violations in creative writing tasks, including word limits, formatting requirements, and content restrictions. For query $q$ and response $o$, the model outputs:

\begin{equation}
V(o, q) = \bigwedge_{i=1}^{n} \text{verify}(o, c_i)
\end{equation}

where $C = \{c_1, c_2, ..., c_n\}$ represents $n$ identified constraints, and $\bigwedge$ denotes logical conjunction. A response is compliant only if all constraints are satisfied.

\subsection{Dynamic Reward Adjustment Strategy}

Fixed-weight reward fusion inadequately balances writing quality and constraint compliance. We introduce a dynamic adjustment mechanism that modifies original rewards before computing GRPO advantages. This ensures constraint-violating samples receive systematic penalties while preserving GRPO's comparative structure.

In standard GRPO, policy $\pi_{\theta_{\text{old}}}$ generates $G$ responses $\{o_1, ..., o_G\}$ for query $q$ with rewards $\{r_1, ..., r_G\}$. Advantages are computed as:

\begin{equation}
\hat{A}_i = \frac{r_i - \text{mean}(\mathbf{r})}{\text{std}(\mathbf{r})}
\end{equation}

Our strategy ensures constraint-violating samples obtain negative advantages after normalization, acting as negative examples during optimization. Compliant samples receive positive advantages and are prioritized for learning.

For each query, we sample $n$ responses $\mathcal{S} = \{s_1, ..., s_n\}$ with original rewards $\{r_1, ..., r_n\}$. We first identify constraint-violating samples through the verification model and adjust their rewards accordingly:

\begin{equation}
r'_i = \begin{cases} 
r_i & \text{if } V(s_i, q) = \text{True} \\
r_i - \delta & \text{if } V(s_i, q) = \text{False} 
\end{cases}
\end{equation}

where $\delta > 0$ is the penalty term to be determined. Let $k$ denote the number of constraint-violating samples in the group. The adjusted mean becomes:

\begin{equation}
\bar{r}' = \frac{1}{n}\sum_{i=1}^{n} r'_i = \frac{1}{n}\left(\sum_{i=1}^{n} r_i - k\delta\right)
\end{equation}

To guarantee that all constraint-violating samples receive negative advantages after normalization, we require that for any violating sample $j$ where $V(s_j, q) = \text{False}$:

\begin{equation}
r'_j < \bar{r}' - \gamma
\end{equation}

where $\gamma > 0$ controls the minimum gap below the adjusted mean. This ensures violating samples will have sufficiently negative advantages to be suppressed during training.

To determine the appropriate penalty $\delta$, let $r_{\text{max}}^{\text{vio}}$ be the highest original reward among all constraint-violating samples. Substituting Equations (4) and (5) into inequality (6), we derive the penalty bound:

\begin{equation}
\delta \geq \frac{n \cdot r_{\text{max}}^{\text{vio}} + n \cdot \gamma - \sum_{i=1}^{n} r_i}{n-k}
\end{equation}

Setting $\delta$ above this bound ensures all violating samples produce negative advantages, systematically suppressing them during gradient updates while preserving the relative ordering among compliant samples. This dynamic adjustment mechanism allows the model to learn from high-quality compliant responses while avoiding the reinforcement of constraint violations.
\subsubsection{Dynamic Sampling Strategy}

Inspired by DAPO~\cite{yu2025dapo}, we address gradient vanishing in creative writing RL training. When all sampled responses receive identical scores, zero advantages yield zero gradients. In creative tasks, this occurs with over-optimized samples, under-optimized samples, and samples where all responses violate constraints.

We implement a composite filtering strategy that removes three types of ineffective samples: (1) groups where all rewards exceed a high threshold, (2) groups where all rewards fall below a low threshold, and (3) groups where all responses fail verification. When filtered samples are insufficient, we dynamically resample new prompts to maintain adequate contrastive signals for effective training.

\section{Experiments and Results}

\begin{table*}[t]
    \centering

    \begin{tabular}{@{}cccccc@{}}
    
    \toprule
    \multirow{2}{*}{\textbf{Model}}               & \multirow{2}{*}{\textbf{Method}}                                                  & \multicolumn{2}{c}{\textbf{Writing Quality}} & \multicolumn{2}{c}{\textbf{Instruction Following}} \\ \cmidrule(l){3-6} 
                                                  &                                                                                   & \textbf{WritingBench}  & \textbf{WriteEval}  & \textbf{ComplexBench}      & \textbf{IFEval}       \\ \midrule
    \multirow{6}{*}{Qwen2.5-32B}                  & Original Model                                                                    & 6.14                   & 3.93\%              & 74.78\%                    & 83.36\%               \\
                                                  & \begin{tabular}[c]{@{}c@{}}GRPO Baseline\\ (Writing RM only)\end{tabular}         & 7.05                   & 7.95\%              & 68.42\%                    & 80.41\%               \\
                                                  & \begin{tabular}[c]{@{}c@{}}GRPO Baseline\\ (Verification Model only)\end{tabular} & 5.73                   & 1.24\%              & \textbf{83.94\%}           & 82.77\%               \\
                                                  & Linear Weighting                                                                  & 7.13                   & 6.40\%              & 73.91\%                    & 84.04\%               \\
                                                  & RLMR(w/o DAPO)                                                                    & 7.34                   & 9.31\%              & 77.83\%                    & \textbf{87.14\%}      \\
                                                  & \textbf{RLMR(Ours)}                                                               & \textbf{7.93}          & \textbf{11.56\%}    & 79.04\%                    & 86.65\%               \\ \midrule
    \multirow{2}{*}{Qwen2.5-72B}                  & Linear Weighting                                                                  & 6.43                   & 10.22\%             & 74.78\%                    & 85.58\%               \\
                                                  & \textbf{RLMR(Ours)}                                                               & \textbf{7.81}          & \textbf{17.18\%}    & \textbf{80.21\%}           & \textbf{87.79\%}      \\ \midrule
    \multirow{2}{*}{Qwen3-8B}                     & Linear Weighting                                                                  & 7.61                   & 26.64\%             & 77.16\%                    & 83.14\%               \\
                                                  & \textbf{RLMR(Ours)}                                                               & \textbf{8.13}          & \textbf{31.69\%}    & \textbf{82.01\%}           & \textbf{86.43\%}      \\ \midrule
    \multirow{2}{*}{DeepSeek-R1-Distill-Llama-8B} & Linear Weighting                                                                  & 5.68                   & 1.46\%              & \textbf{53.91\%}           & 56.38\%               \\
                                                  & \textbf{RLMR(Ours)}                                                               & \textbf{7.41}          & \textbf{3.57\%}     & 52.35\%                    & \textbf{60.94\%}      \\ \bottomrule
    \end{tabular}
    \caption{Performance comparison across different models and methods on writing quality and instruction-following benchmarks. Our dynamic mixed-reward approach consistently outperforms baseline methods across all model scales.}
    \label{tab:main_results}
    \end{table*}

\begin{figure*}[t]
\centering
\begin{subfigure}[t]{0.33\textwidth}
\includegraphics[width=\textwidth]{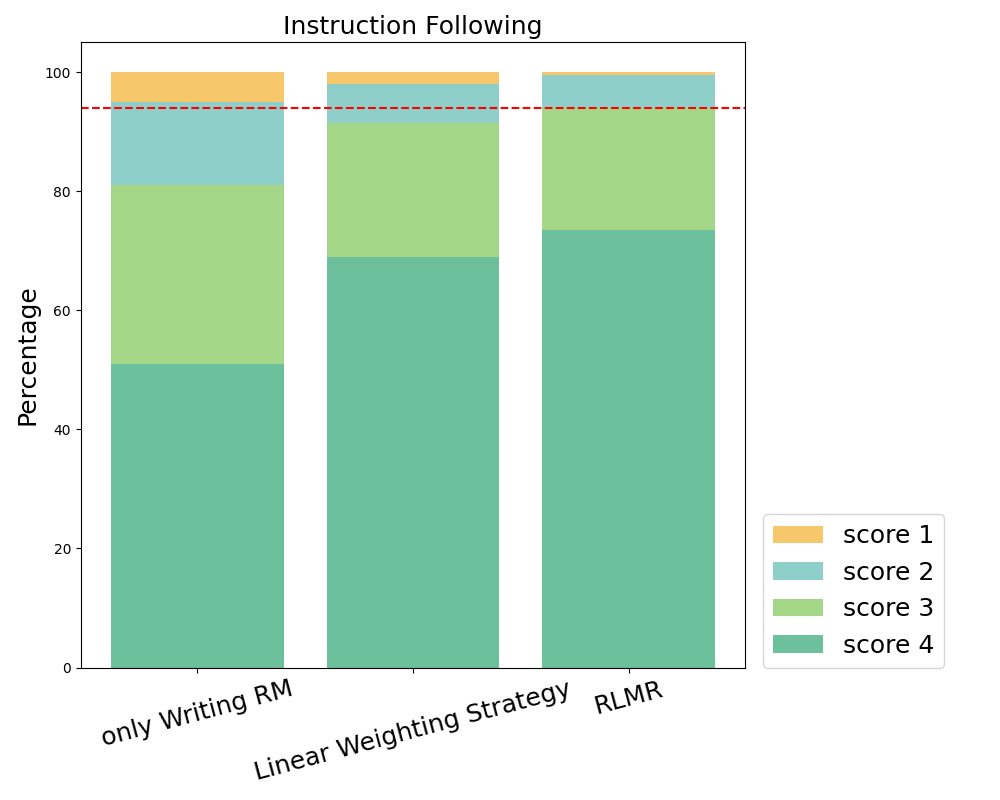}
\caption{Instruction Following}
\label{fig:instruction_following}
\end{subfigure}
\hfill
\begin{subfigure}[t]{0.33\textwidth}
\includegraphics[width=\textwidth]{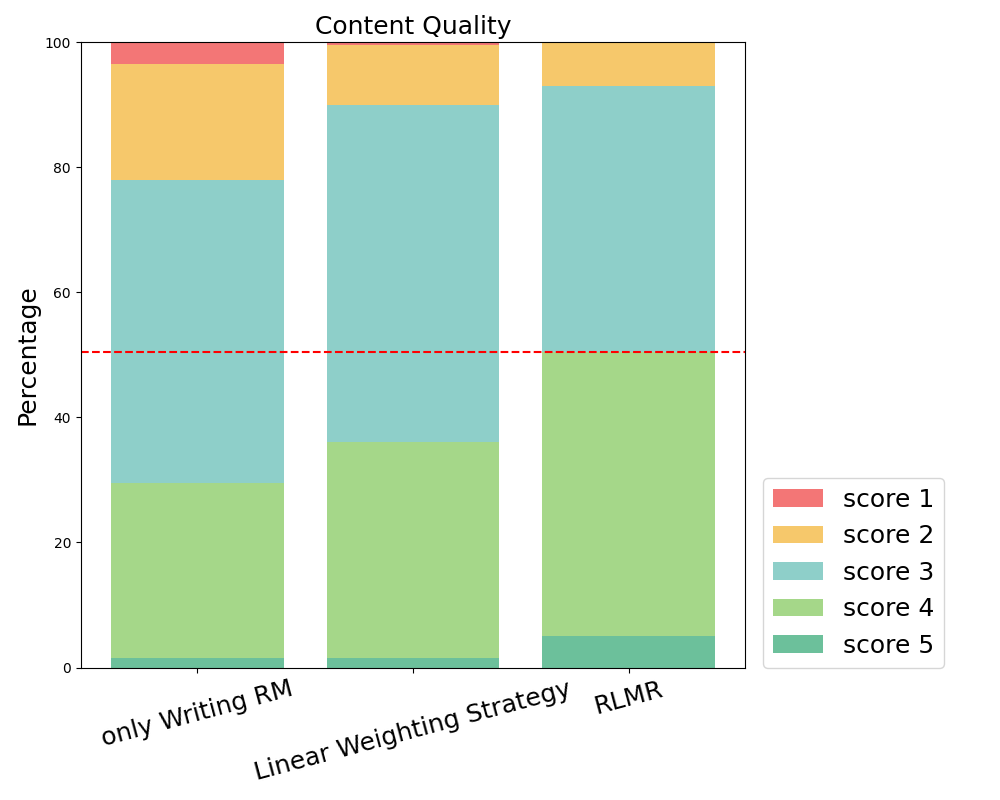}
\caption{Content Quality}
\label{fig:content_quality}
\end{subfigure}
\hfill
\begin{subfigure}[t]{0.33\textwidth}
\includegraphics[width=\textwidth]{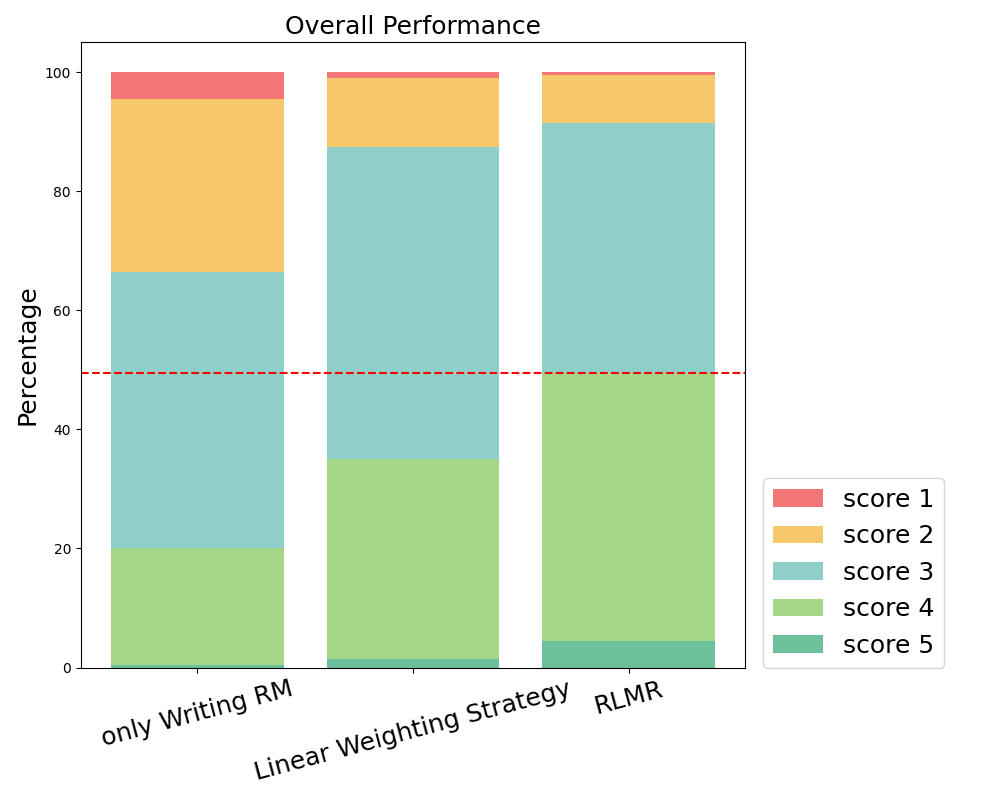}
\caption{Overall Performance}
\label{fig:overall_performance}
\end{subfigure}
\caption{Human evaluation score distributions across three dimensions. The red dashed line indicates the satisfactory threshold (score $\geq$ 3 for Content Quality and Overall Performance, score = 4 for Instruction Following). Our RLMR method consistently shows higher proportions of satisfactory scores compared to baseline methods.}
\label{fig:human_eval_scores}
\end{figure*}

In this section, we show experiments to test our dynamic mixed-reward GRPO framework for creative writing. We describe the setup, share results, and give analysis.

\subsection{Experimental Setup}
\label{sec:exp_config}
\subsubsection{Training Query Construction}

We construct our GRPO training queries from real-world seed data, we apply the self-instruct~\cite{wang2023selfinstructaligninglanguagemodels} methodology to expand the dataset diversity while maintaining realistic writing scenarios. To ensure balanced genre representation, we employ DeepSeek-V3 to classify generated queries by writing genre and adjust the sampling distribution to match real-world proportions observed in our seed data. This process yields a final training set of 8,739 queries.

\subsubsection{Evaluation Benchmarks}

We test model performance on writing quality and instruction following using four benchmarks:

\textbf{WritingBench}~\cite{writingbench2025} covers 6 main categories and 100 subdomains like academic, finance, politics, literature, education, and marketing. It has 1,239 real-world prompts, each with 5 custom criteria. We use Claude-4-Sonnet to score outputs.

\textbf{WriteEval} is our custom dataset containing 890 samples collected from real-world scenarios and augmented with LLM-generated instructions to match authentic writing styles. The dataset uniformly covers 30 primary writing genres and 377 secondary categories, including Chinese-specific genres such as folk texts, classical Chinese, and composition writing. For each instruction, we solicited responses from six competitive Chinese writing models: Claude-4-Sonnet, Gemini-2.5-Pro, DeepSeek-R1, DeepSeek-V3, Doubao-1.5-Thinking, and Hunyuan-TurboS. Human experts conducted blind evaluation to select the best response from each set as reference answers. For automated evaluation, Claude-4-Opus compares model outputs against reference answers to determine win rates:
$\text{Win Rate} = \frac{\text{Number of wins}}{\text{Total comparisons}} \times 100\%$
where a "win" indicates the model output is judged superior to the reference answer. Detailed prompt templates are provided in the appendix.

\textbf{ComplexBench}~\cite{wen2024benchmarking} checks complex instruction following with combined constraints. It builds hard prompts that need to meet multiple rules. Scoring uses questions to check each part.

\textbf{IFEval}~\cite{zhou2023instructionfollowingevaluationlargelanguage} is Google's benchmark for verifiable instructions like word count or keywords. It has 25 types across 500 prompts. We use prompt-level strict-accuracy for evaluation.

\subsubsection{Baseline Methods}

To evaluate our dynamic mixed-reward strategy, we compare against three baseline methods that represent the spectrum of existing reward strategies in creative writing optimization:

\textbf{(1) Writing Reward Only GRPO}: This baseline trains using only writing quality rewards without any constraint verification signals. This method represents the traditional approach in RLHF where models are optimized solely based on human preference signals for output quality~\cite{ouyang2022training,stiennon2020learning}. Following established RLHF practices, this baseline uses a reward model trained on human-annotated preference pairs to score creative writing outputs~\cite{dong2024rlhf}.

\textbf{(2) Verification Signal Only GRPO}: This baseline uses only binary constraint verification signals (pass/fail) without considering writing quality. This approach aligns with recent work on Reinforcement Learning with Verifiable Rewards (RLVR), where models are trained using deterministic verification functions for tasks with clear correctness criteria~\cite{cobbe2021training,mroueh2025reinforcement}. TheBy comparing against these methods, we demonstrate that our dynamic mixed-reward strategy addresses the limitations of both single-reward and fixed-weight approaches, providing a more effective solution for creative writing optimization.

\textbf{(3) Linear Weighting Strategy}: Following the approach proposed by~\citet{peng-etal-2025-agentic}, this baseline combines writing rewards with verification signals through fixed-weight linear combination. Specifically, we normalize both writing rewards and verification scores to the [0,1] range and compute their arithmetic mean: $(s_{\text{normalized\_writing}} + s_{\text{normalized\_verification}}) / 2$. This method represents the current state-of-the-art in mixed-reward strategies, as demonstrated in the Agentic Reward Modeling framework~\cite{peng-etal-2025-agentic}, which successfully integrates human preference rewards with verifiable correctness signals including factuality and instruction following.

\subsubsection{Reward Model and Training Setup}

\paragraph{Writing Reward Model.} 
We use a Pointwise Bradley-Terry Reward Model~\cite{bradley1952rank,ouyang2022training} for continuous feedback. It trains on Tencent-Hunyuan-Large~\cite{sun2024hunyuanlargeopensourcemoemodel} with 200,000 labeled samples. Each sample has a prompt and two responses; humans pick the better one based on quality, adherence, style, and experience. We use this model for rewards in RLHF to match human preferences.

\paragraph{Constraint Verification Model.}
We use 
Qwen2.5-72B-Instruct  with prompts to check constraints. It makes checklists and verifies each one. We employ binary verification (all constraints satisfied or not) rather than proportion-based scoring because partial constraint satisfaction is functionally equivalent to complete failure in creative writing tasks. This binary approach ensures the model learns to generate responses that satisfy all constraints simultaneously, rather than trading off between different constraint types. See appendix for prompt details.

\begin{figure}[h!]
\centering
\includegraphics[width=0.5\textwidth]{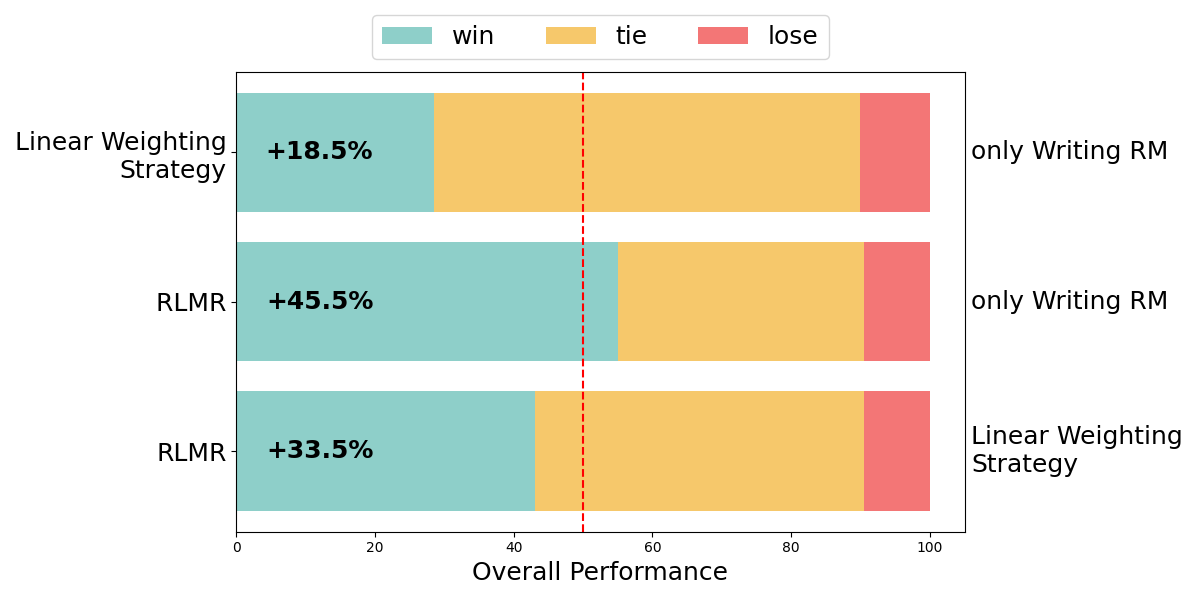}
\caption{Pairwise comparison results for Overall Performance. "Win" indicates the left method outperforms the right method; "tie" indicates comparable performance; "lose" indicates the left method underperforms. The red dashed line represents equal performance (50\%). RLMR demonstrates significant advantages over both baseline methods.}
\label{fig:pairwise_comparison}
\end{figure}
\begin{figure*}[ht]
\centering
\begin{subfigure}[t]{0.32\textwidth}
\includegraphics[width=\textwidth]{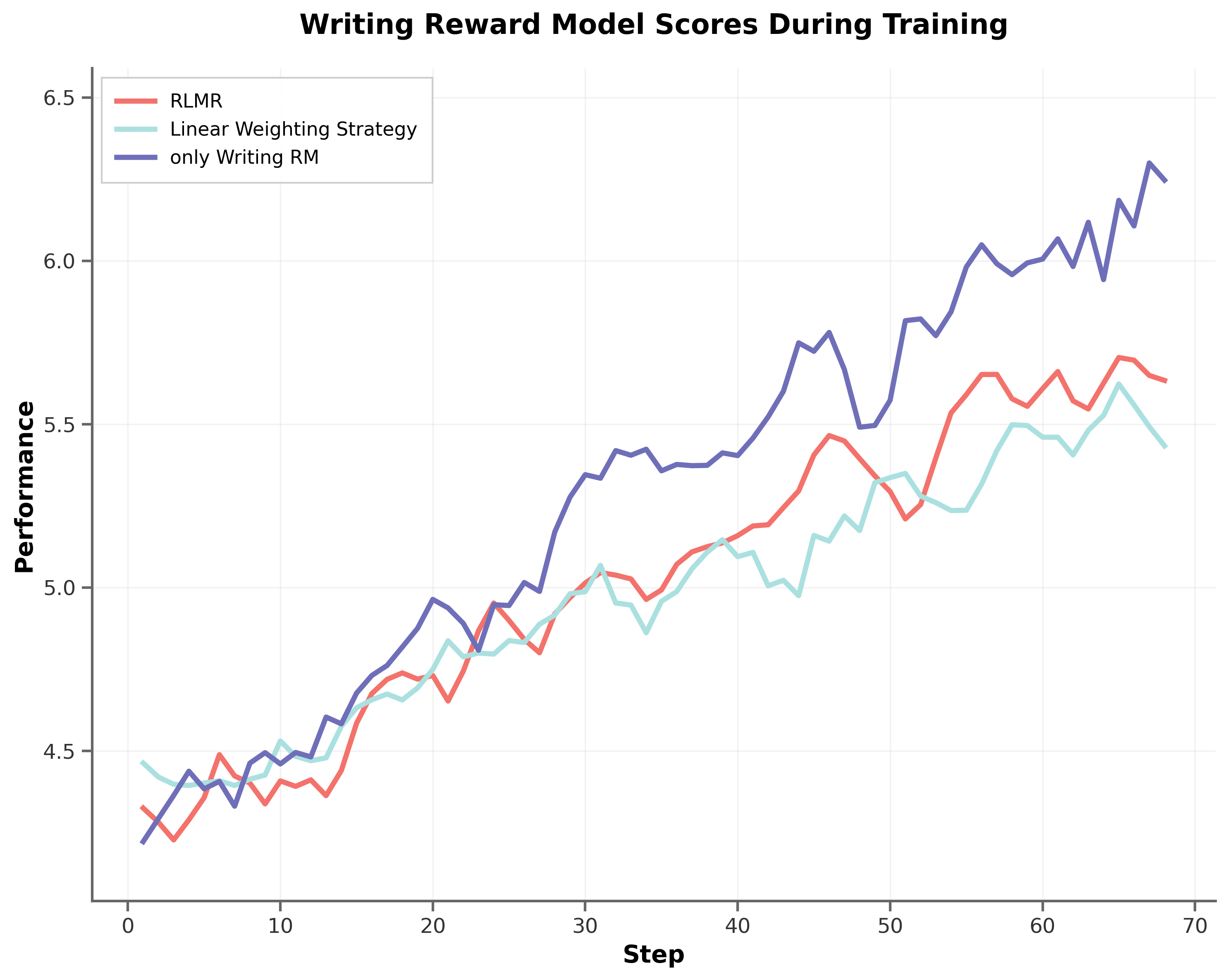}
\caption{Writing Reward Scores}
\label{fig:reward_scores}
\end{subfigure}
\hfill
\begin{subfigure}[t]{0.32\textwidth}
\includegraphics[width=\textwidth]{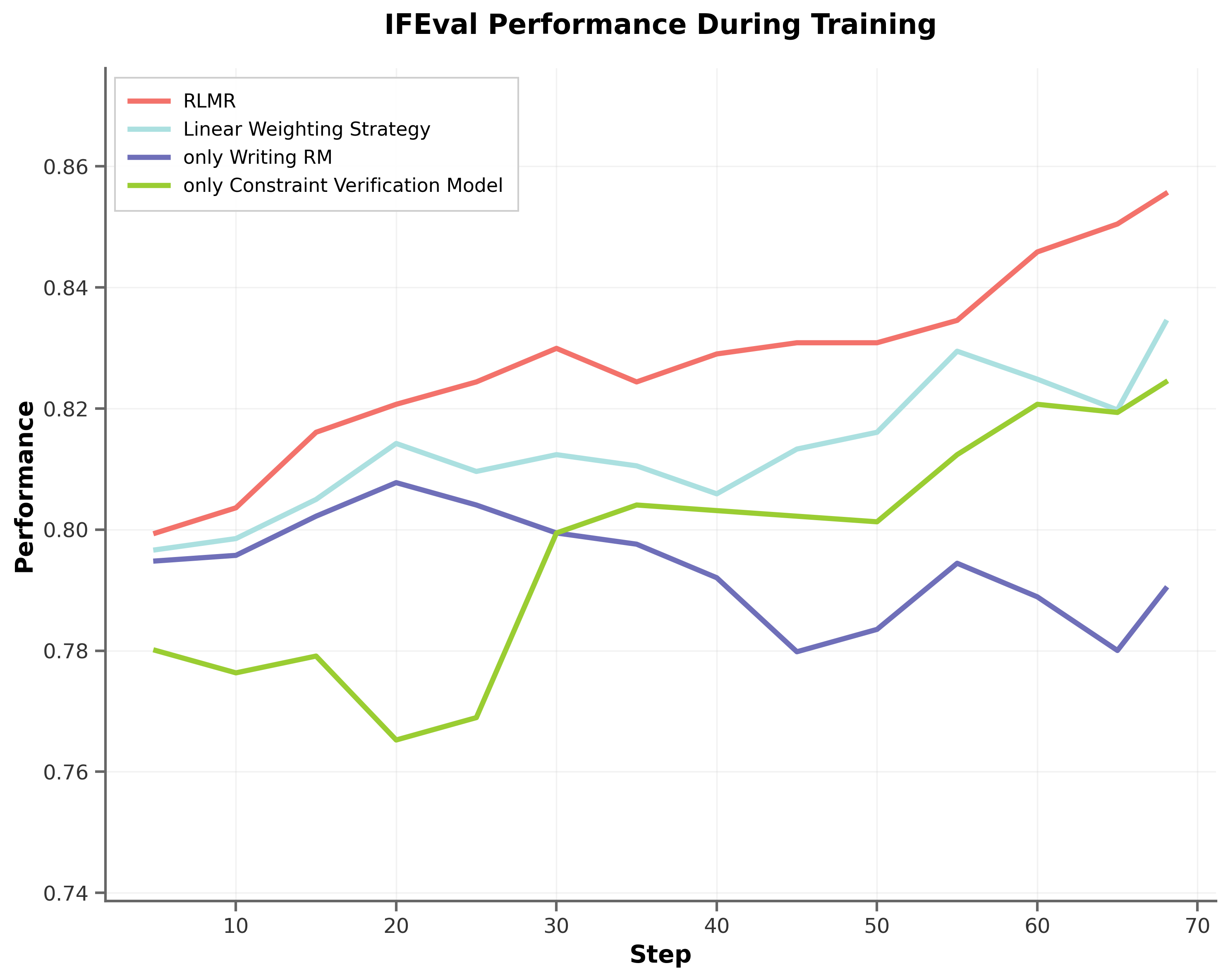}
\caption{IFEval Performance}
\label{fig:ifeval_performance}
\end{subfigure}
\hfill
\begin{subfigure}[t]{0.32\textwidth}
\includegraphics[width=\textwidth]{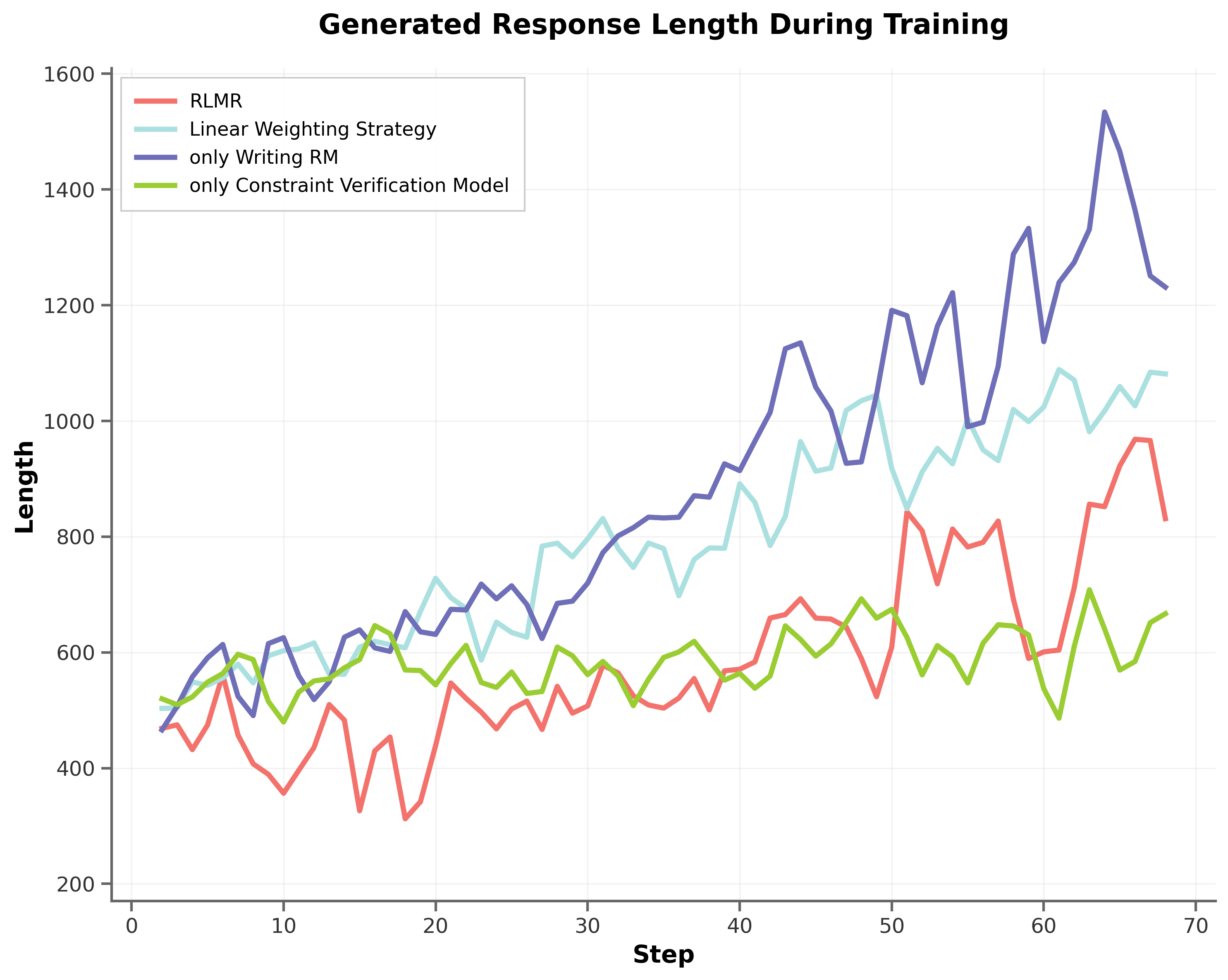}
\caption{Response Length}
\label{fig:response_length}
\end{subfigure}
\caption{Training dynamics across different metrics. (a) Writing reward model scores during training. (b) IFEval performance during training. (c) Generated response length during training.}
\label{fig:training_dynamics}
\end{figure*}

\subsection{Experimental Results}

\subsubsection{Automated Evaluation Results}

We test our framework on four models: Qwen2.5-32B and Qwen2.5-72B~\cite{qwen2.5,qwen2}, Qwen3-8B~\cite{qwen3}, and DeepSeek-R1-Distill-Llama-8B~\cite{deepseekai2025deepseekr1incentivizingreasoningcapability}. Table~\ref{tab:main_results} shows results across methods and benchmarks.

The automated evaluation results reveal compelling evidence for the effectiveness of our dynamic mixed-reward approach. Results from Qwen2.5-32B clearly expose the inherent problems with single reward signals. When training with writing RM alone, writing quality improves modestly from 6.14 to 6.35, yet instruction following suffers substantial degradation: ComplexBench performance drops from 74.78\% to 68.42\%, while IFEval accuracy falls from 83.36\% to 80.41\%. The reverse pattern emerges when using only the constraint verification model—instruction following on ComplexBench rises from 74.78\% to 83.94\%, but writing quality plummets from 6.14 to 5.73, with WriteEval performance collapsing from 3.93\% to a mere 1.24\%. This stark trade-off demonstrates that single signals cannot balance subjective creative quality with objective constraint adherence.

Given these limitations, mixed-reward strategies emerge as a natural solution by combining writing RM with constraint verification signals. The most classical approach is linear weighting, which averages the two reward types with fixed coefficients. On Qwen2.5-32B, this approach elevates writing quality to 7.13 while preserving reasonable instruction following capabilities, successfully avoiding the severe bias problems observed with single-signal methods. These results underscore the critical importance of integrating both subjective and objective evaluation dimensions in creative writing optimization.

However, our RLMR method delivers even greater improvements, consistently outperforming linear weighting across all tested models. On Qwen2.5-32B, RLMR pushes writing quality further to 7.93 and achieves an 11.56\% WriteEval win rate, substantially surpassing linear weighting's 7.13 and 6.40\% respectively. This pattern of superior performance extends to other architectures: Qwen3-8B sees writing quality rise from 7.61 to 8.13, with WriteEval win rates jumping from 26.64\% to 31.69\%. Similarly, Qwen2.5-72B confirms this trend, with WriteEval performance climbing from 10.22\% to 17.18\%.

The robustness of these improvements becomes evident when examining results across diverse model scales. Our experiments span architectures ranging from 8B to 72B parameters, including both Qwen and DeepSeek families, with all models demonstrating consistent advantages under RLMR. 
\subsubsection{Manual Evaluation Results}

We conducted human evaluation on 200 randomly sampled instances from the WriteEval dataset to assess model performance across three dimensions: Instruction Following, Content Quality, and Overall Performance. Detailed scoring criteria and guidelines are provided in the appendix. For Instruction Following, we consider a score of 4 as complete instruction adherence. For Content Quality and Overall Performance, scores of 3 or above are considered satisfactory.

Figure~\ref{fig:human_eval_scores} presents the score distribution across all three evaluation dimensions. The results clearly demonstrate the limitations of single-reward strategies. The writing-only baseline shows inferior performance across multiple dimensions compared to mixed-reward approaches, with notably lower satisfactory rates in instruction following and content quality. Among mixed-reward strategies, our RLMR method achieves higher satisfactory rates across all dimensions.

Specifically, for Instruction Following, RLMR shows the highest proportion of perfect scores (score 4), indicating superior constraint adherence. In Content Quality, RLMR demonstrates a more favorable distribution with increased proportions in higher score ranges (scores 4-5), suggesting better content generation capabilities. The Overall Performance dimension reveals similar trends, with RLMR achieving the most balanced distribution toward higher satisfaction levels.

Figure~\ref{fig:pairwise_comparison} shows the results of direct pairwise comparisons for Overall Performance. RLMR achieves substantial win rates against both baseline methods: 45.5\% win rate versus writing-only baseline and 33.5\% win rate versus linear weighting strategy. These results demonstrate that our RLMR strategy achieves higher usability and satisfaction rates in creative writing tasks, confirming the practical effectiveness of our approach.

\subsection{Experimental Analysis}

The experimental results demonstrate that single reward signals fail to balance writing quality and instruction following effectively. Using only writing rewards improves creative quality but reduces constraint adherence. Using only verification signals severely harms writing quality while providing limited gains in instruction following. These findings confirm that creative writing optimization requires careful integration of both subjective and objective evaluation criteria.

Our dynamic mixed-reward strategy significantly outperforms linear weighting approaches across all tested models and benchmarks. This superiority stems from fundamental limitations of fixed-weight methods. Writing quality scores and constraint verification signals operate on different scales and distributions. Writing rewards typically follow continuous distributions, while constraint verification produces binary outcomes. The scalar inconsistency between these two signals makes it difficult to determine appropriate weighting coefficients. Moreover, optimal weighting coefficients need adjustment for different reward models, making fixed-weight approaches impractical across diverse model configurations.

Our dynamic adjustment mechanism addresses these limitations by calculating penalty terms based on actual constraint compliance patterns within each sample group. Rather than applying uniform weights, the approach modulates penalties according to the theoretical bounds derived in Equation (7). This ensures constraint-violating samples consistently receive negative advantages and are suppressed during training.

Figure~\ref{fig:training_dynamics} shows training dynamics across key metrics. The writing RM only baseline achieves the highest writing reward scores during training (Figure~\ref{fig:reward_scores}), but this improvement reveals classic reward hacking behavior. Despite high reward scores, its IFEval performance deteriorates significantly (Figure~\ref{fig:ifeval_performance}), dropping below both the original model and other baselines. This divergence between reward scores and actual instruction-following capability demonstrates that the model learns to exploit the reward model rather than genuinely improving writing quality.

The reward hacking behavior is further evidenced by the dramatic increase in response length (Figure~\ref{fig:response_length}). The writing RM only baseline shows uncontrolled length growth, reaching over 1400 tokens on average, which explains its poor instruction-following performance. When models generate excessively long outputs, they cannot properly adhere to specific constraints like word count limits, format requirements, or conciseness instructions.

In contrast, our RLMR method maintains balanced optimization across all metrics. It achieves steady improvement in writing reward scores while preserving strong IFEval performance, demonstrating that our dynamic reward adjustment successfully prevents the model from exploiting either reward signal. The controlled response length further confirms that RLMR learns to generate high-quality content without resorting to length inflation. This balanced training dynamic validates the effectiveness of our dynamic penalty mechanism in creating models that excel at both creative quality and constraint adherence.
\section{Conclusion}

 we proposed RLMR (Reinforcement Learning with Mixed Rewards), a dynamic mixed-reward GRPO framework that addresses the fundamental challenge of balancing subjective creative quality with objective constraint adherence in creative writing optimization. By developing a dynamic reward adjustment mechanism that ensures constraint-violating samples receive negative advantages during training, our method overcomes the limitations of both single-reward and fixed-weight strategies. Experimental results across diverse model architectures demonstrate that RLMR achieves substantial improvements in both writing quality (11.56\% WriteEval win rate on Qwen2.5-32B) and constraint compliance (86.65\% IFEval accuracy), with human evaluation confirming significant user preference. The training dynamics analysis reveals that our method successfully prevents reward hacking while maintaining stable optimization, providing a principled and computationally efficient solution to multi-objective creative writing optimization. Future work includes extending this framework to other multi-signal scenarios such as dialogue systems and code generation.

\bibliography{aaai2026}

\begin{thebibliography}{22}
\providecommand{\natexlab}[1]{#1}

\bibitem[{Bradley and Terry(1952)}]{bradley1952rank}
Bradley, R.~A.; and Terry, M.~E. 1952.
\newblock Rank analysis of incomplete block designs: I. the method of paired comparisons.
\newblock \emph{Biometrika}, 39(3/4): 324--345.

\bibitem[{Cobbe et~al.(2021)Cobbe, Kosaraju, Bavarian, Chen, Jun, Kaiser, Plappert, Tworek, Hilton, Nakano et~al.}]{cobbe2021training}
Cobbe, K.; Kosaraju, V.; Bavarian, M.; Chen, M.; Jun, H.; Kaiser, L.; Plappert, M.; Tworek, J.; Hilton, J.; Nakano, R.; et~al. 2021.
\newblock Training verifiers to solve math word problems.
\newblock \emph{arXiv preprint arXiv:2110.14168}.

\bibitem[{DeepSeek-AI(2025)}]{deepseekai2025deepseekr1incentivizingreasoningcapability}
DeepSeek-AI. 2025.
\newblock DeepSeek-R1: Incentivizing Reasoning Capability in LLMs via Reinforcement Learning.
\newblock arXiv:2501.12948.

\bibitem[{Dong et~al.(2024)Dong, Xiong, Pang, Wang, Zhao, Zhou, Jiang, Sahoo, Xiong, and Zhang}]{dong2024rlhf}
Dong, H.; Xiong, W.; Pang, B.; Wang, H.; Zhao, H.; Zhou, Y.; Jiang, N.; Sahoo, D.; Xiong, C.; and Zhang, T. 2024.
\newblock RLHF Workflow: From Reward Modeling to Online RLHF.
\newblock \emph{arXiv preprint arXiv:2405.07863}.

\bibitem[{Jia et~al.(2025)Jia, Yang, Gai, Luo, Huang, Lin, Jiang, and Jiang}]{jia2025writingzero}
Jia, R.; Yang, Y.; Gai, Y.; Luo, K.; Huang, S.; Lin, J.; Jiang, X.; and Jiang, G. 2025.
\newblock Writing-Zero: Bridge the Gap Between Non-verifiable Tasks and Verifiable Rewards.
\newblock \emph{arXiv preprint arXiv:2506.00103}.

\bibitem[{Mroueh(2025)}]{mroueh2025reinforcement}
Mroueh, Y. 2025.
\newblock Reinforcement Learning with Verifiable Rewards: GRPO's Effective Loss, Dynamics, and Success Amplification.
\newblock \emph{arXiv preprint arXiv:2503.06639}.

\bibitem[{Ouyang et~al.(2022)Ouyang, Wu, Jiang et~al.}]{ouyang2022training}
Ouyang, L.; Wu, J.; Jiang, X.; et~al. 2022.
\newblock Training language models to follow instructions with human feedback.
\newblock \emph{Advances in Neural Information Processing Systems}, 35: 27730--27744.

\bibitem[{Peng et~al.(2025{\natexlab{a}})Peng, Qi, Wang, Yao, Xu, Hou, and Li}]{peng2025agentic}
Peng, H.; Qi, Y.; Wang, X.; Yao, Z.; Xu, B.; Hou, L.; and Li, J. 2025{\natexlab{a}}.
\newblock Agentic Reward Modeling: Integrating Human Preferences with Verifiable Correctness Signals for Reliable Reward Systems.
\newblock In Che, W.; Nabende, J.; Shutova, E.; and Pilehvar, M.~T., eds., \emph{Proceedings of the 63rd Annual Meeting of the Association for Computational Linguistics (Volume 1: Long Papers)}, 15934--15949. Vienna, Austria: Association for Computational Linguistics.
\newblock ISBN 979-8-89176-251-0.

\bibitem[{Peng et~al.(2025{\natexlab{b}})Peng, Qi, Wang, Yao, Xu, Hou, and Li}]{peng-etal-2025-agentic}
Peng, H.; Qi, Y.; Wang, X.; Yao, Z.; Xu, B.; Hou, L.; and Li, J. 2025{\natexlab{b}}.
\newblock Agentic Reward Modeling: Integrating Human Preferences with Verifiable Correctness Signals for Reliable Reward Systems.
\newblock In Che, W.; Nabende, J.; Shutova, E.; and Pilehvar, M.~T., eds., \emph{Proceedings of the 63rd Annual Meeting of the Association for Computational Linguistics (Volume 1: Long Papers)}, 15934--15949. Vienna, Austria: Association for Computational Linguistics.
\newblock ISBN 979-8-89176-251-0.

\bibitem[{Shao et~al.(2024)Shao, Wang, Zhu, Xu, Song, Bi, Zhang, Zhang, Li, Wu, and Guo}]{shao2024deepseekmath}
Shao, Z.; Wang, P.; Zhu, Q.; Xu, R.; Song, J.; Bi, X.; Zhang, H.; Zhang, M.; Li, Y.; Wu, Y.; and Guo, D. 2024.
\newblock DeepSeekMath: Pushing the Limits of Mathematical Reasoning in Open Language Models.
\newblock \emph{arXiv preprint arXiv:2402.03300}.

\bibitem[{Sheng et~al.(2024)Sheng, Zhang, Ye, Wu, Zhang, Zhang, Peng, Lin, and Wu}]{sheng2024hybridflow}
Sheng, G.; Zhang, C.; Ye, Z.; Wu, X.; Zhang, W.; Zhang, R.; Peng, Y.; Lin, H.; and Wu, C. 2024.
\newblock HybridFlow: A Flexible and Efficient RLHF Framework.
\newblock \emph{arXiv preprint arXiv: 2409.19256}.

\bibitem[{Stiennon et~al.(2020)Stiennon, Ouyang, Wu, Ziegler, Lowe, Voss, Radford, Amodei, and Christiano}]{stiennon2020learning}
Stiennon, N.; Ouyang, L.; Wu, J.; Ziegler, D.; Lowe, R.; Voss, C.; Radford, A.; Amodei, D.; and Christiano, P.~F. 2020.
\newblock Learning to summarize with human feedback.
\newblock \emph{Advances in Neural Information Processing Systems}, 33: 3008--3021.

\bibitem[{Sun et~al.(2024)Sun, Chen, Huang, Xie, Zhu, Zhang, Li, Yang, Han, Shu, Bu, Chen, Huang, Lian, Yang, Yan, Zeng, Ren, Yu, Wu, Mao, Xia, Yang, Zheng, Wu, Jiao, Xue, Zhang, Wu, Liu, Wu, Xu, Chen, Chen, Feng, Hong, Zheng, Xu, Li, Kuang, Hu, Chen, Deng, Li, Liu, Zhang, Hu, Zhao, Wu, Ding, Wang, Liu, Wang, Fei, Yu, Zhao, Cao, Wang, Xiang, Huang, Xiong, Hu, Hou, Jiang, Ma, Wu, Deng, Shen, Wang, Liu, Liu, Chen, Dong, Jia, Chen, Liu, Yuan, Xu, Yan, Cao, Hu, Feng, Du, Yu, Tao, Zhang, Zhu, Xu, Li, Zha, Ouyang, Xia, Li, He, Chen, Song, Chen, Jiang, Zhao, Wang, Gong, Gan, Hu, Kang, Yang, Liu, Wang, and Jiang}]{sun2024hunyuanlargeopensourcemoemodel}
Sun, X.; Chen, Y.; Huang, Y.; Xie, R.; Zhu, J.; Zhang, K.; Li, S.; Yang, Z.; Han, J.; Shu, X.; Bu, J.; Chen, Z.; Huang, X.; Lian, F.; Yang, S.; Yan, J.; Zeng, Y.; Ren, X.; Yu, C.; Wu, L.; Mao, Y.; Xia, J.; Yang, T.; Zheng, S.; Wu, K.; Jiao, D.; Xue, J.; Zhang, X.; Wu, D.; Liu, K.; Wu, D.; Xu, G.; Chen, S.; Chen, S.; Feng, X.; Hong, Y.; Zheng, J.; Xu, C.; Li, Z.; Kuang, X.; Hu, J.; Chen, Y.; Deng, Y.; Li, G.; Liu, A.; Zhang, C.; Hu, S.; Zhao, Z.; Wu, Z.; Ding, Y.; Wang, W.; Liu, H.; Wang, R.; Fei, H.; Yu, P.; Zhao, Z.; Cao, X.; Wang, H.; Xiang, F.; Huang, M.; Xiong, Z.; Hu, B.; Hou, X.; Jiang, L.; Ma, J.; Wu, J.; Deng, Y.; Shen, Y.; Wang, Q.; Liu, W.; Liu, J.; Chen, M.; Dong, L.; Jia, W.; Chen, H.; Liu, F.; Yuan, R.; Xu, H.; Yan, Z.; Cao, T.; Hu, Z.; Feng, X.; Du, D.; Yu, T.; Tao, Y.; Zhang, F.; Zhu, J.; Xu, C.; Li, X.; Zha, C.; Ouyang, W.; Xia, Y.; Li, X.; He, Z.; Chen, R.; Song, J.; Chen, R.; Jiang, F.; Zhao, C.; Wang, B.; Gong, H.; Gan, R.; Hu, W.; Kang, Z.; Yang, Y.; Liu, Y.; Wang, D.; and Jiang, J. 2024.
\newblock Hunyuan-Large: An Open-Source MoE Model with 52 Billion Activated Parameters by Tencent.
\newblock arXiv:2411.02265.

\bibitem[{Team(2024)}]{qwen2.5}
Team, Q. 2024.
\newblock Qwen2.5: A Party of Foundation Models.

\bibitem[{Wang et~al.(2023)Wang, Kordi, Mishra, Liu, Smith, Khashabi, and Hajishirzi}]{wang2023selfinstructaligninglanguagemodels}
Wang, Y.; Kordi, Y.; Mishra, S.; Liu, A.; Smith, N.~A.; Khashabi, D.; and Hajishirzi, H. 2023.
\newblock Self-Instruct: Aligning Language Models with Self-Generated Instructions.
\newblock arXiv:2212.10560.

\bibitem[{Wen et~al.(2024)Wen, Ke, Gu, Wu, Huang, Zhou, Li, Hu, Gao, Xu et~al.}]{wen2024benchmarking}
Wen, B.; Ke, P.; Gu, X.; Wu, L.; Huang, H.; Zhou, J.; Li, W.; Hu, B.; Gao, W.; Xu, J.; et~al. 2024.
\newblock Benchmarking Complex Instruction-Following with Multiple Constraints Composition.
\newblock \emph{arXiv preprint arXiv:2407.03978}.

\bibitem[{Wu et~al.(2025)Wu, Bai, Hu, Lee, and Li}]{wu2025longwriter}
Wu, Y.; Bai, Y.; Hu, Z.; Lee, R. K.-W.; and Li, J. 2025.
\newblock LongWriter-Zero: Mastering Ultra-Long Text Generation via Reinforcement Learning.
\newblock \emph{arXiv preprint arXiv:2506.18841}.

\bibitem[{Yang et~al.(2024)Yang, Yang, Hui, Zheng, Yu, Zhou, Li, Li, Liu, Huang, Dong, Wei, Lin, Tang, Wang, Yang, Tu, Zhang, Ma, Xu, Zhou, Bai, He, Lin, Dang, Lu, Chen, Yang, Li, Xue, Ni, Zhang, Wang, Peng, Men, Gao, Lin, Wang, Bai, Tan, Zhu, Li, Liu, Ge, Deng, Zhou, Ren, Zhang, Wei, Ren, Fan, Yao, Zhang, Wan, Chu, Liu, Cui, Zhang, and Fan}]{qwen2}
Yang, A.; Yang, B.; Hui, B.; Zheng, B.; Yu, B.; Zhou, C.; Li, C.; Li, C.; Liu, D.; Huang, F.; Dong, G.; Wei, H.; Lin, H.; Tang, J.; Wang, J.; Yang, J.; Tu, J.; Zhang, J.; Ma, J.; Xu, J.; Zhou, J.; Bai, J.; He, J.; Lin, J.; Dang, K.; Lu, K.; Chen, K.; Yang, K.; Li, M.; Xue, M.; Ni, N.; Zhang, P.; Wang, P.; Peng, R.; Men, R.; Gao, R.; Lin, R.; Wang, S.; Bai, S.; Tan, S.; Zhu, T.; Li, T.; Liu, T.; Ge, W.; Deng, X.; Zhou, X.; Ren, X.; Zhang, X.; Wei, X.; Ren, X.; Fan, Y.; Yao, Y.; Zhang, Y.; Wan, Y.; Chu, Y.; Liu, Y.; Cui, Z.; Zhang, Z.; and Fan, Z. 2024.
\newblock Qwen2 Technical Report.
\newblock \emph{arXiv preprint arXiv:2407.21783}.

\bibitem[{Yang et~al.(2025)Yang, Yang, Hui, Zheng, Yu, Zhou, Li, Li, Liu, Huang, Dong, Wei, Lin, Tang, Wang, Yang, Tu, Zhang, Ma, Xu, Zhou, Bai, He, Lin, Dang, Lu, Chen, Yang, Li, Xue, Ni, Zhang, Wang, Peng, Men, Gao, Lin, Wang, Bai, Tan, Zhu, Li, Liu, Ge, Deng, Zhou, Ren, Zhang, Wei, Ren, Fan, Yao, Zhang, Wan, Chu, Liu, Cui, Zhang, and Fan}]{qwen3}
Yang, A.; Yang, B.; Hui, B.; Zheng, B.; Yu, B.; Zhou, C.; Li, C.; Li, C.; Liu, D.; Huang, F.; Dong, G.; Wei, H.; Lin, H.; Tang, J.; Wang, J.; Yang, J.; Tu, J.; Zhang, J.; Ma, J.; Xu, J.; Zhou, J.; Bai, J.; He, J.; Lin, J.; Dang, K.; Lu, K.; Chen, K.; Yang, K.; Li, M.; Xue, M.; Ni, N.; Zhang, P.; Wang, P.; Peng, R.; Men, R.; Gao, R.; Lin, R.; Wang, S.; Bai, S.; Tan, S.; Zhu, T.; Li, T.; Liu, T.; Ge, W.; Deng, X.; Zhou, X.; Ren, X.; Zhang, X.; Wei, X.; Ren, X.; Fan, Y.; Yao, Y.; Zhang, Y.; Wan, Y.; Chu, Y.; Liu, Y.; Cui, Z.; Zhang, Z.; and Fan, Z. 2025.
\newblock Qwen3 Technical Report.
\newblock \emph{arXiv preprint arXiv:2505.09388}.

\bibitem[{Yao et~al.(2025)}]{writingbench2025}
Yao, L.; et~al. 2025.
\newblock WritingBench: A Comprehensive Benchmark for Generative Writing.
\newblock \emph{arXiv preprint arXiv:2503.05244}.

\bibitem[{Yu et~al.(2025)Yu, Liu, Chen et~al.}]{yu2025dapo}
Yu, Y.; Liu, Y.; Chen, H.; et~al. 2025.
\newblock DAPO: An Open-Source LLM Reinforcement Learning System at Scale.
\newblock \emph{arXiv preprint arXiv:2503.14476}.

\bibitem[{Zhou et~al.(2023)Zhou, Lu, Mishra, Brahma, Basu, Luan, Zhou, and Hou}]{zhou2023instructionfollowingevaluationlargelanguage}
Zhou, J.; Lu, T.; Mishra, S.; Brahma, S.; Basu, S.; Luan, Y.; Zhou, D.; and Hou, L. 2023.
\newblock Instruction-Following Evaluation for Large Language Models.
\newblock \emph{arXiv preprint arXiv:2311.07911}.

\end{thebibliography}

\clearpage
\section{Appendix}

\subsection{Manual Evaluation Criteria}

This section describes the evaluation criteria used to assess AI-generated responses in our human evaluation study. 

Our evaluation uses three distinct dimensions to capture different aspects of response quality: \textbf{Instruction Following}, \textbf{Content Quality}, and \textbf{Overall Performance}. Each dimension focuses on specific characteristics that together provide comprehensive coverage of response effectiveness.The scoring criteria are shown in Table~\ref{tab:scoring_rubric}.
\begin{table*}
\centering
\caption{Response Quality Scoring Rubric}
\label{tab:scoring_rubric}
\footnotesize
\begin{tabular}{|c|p{4.5cm}|p{4.5cm}|p{4.5cm}|}
\hline
\textbf{Score} & \textbf{Instruction Following} & \textbf{Content Quality} & \textbf{Overall Performance} \\
\hline
\textbf{1} & 
Complete misunderstanding of user intent. Fails to address core requirements. Produces wrong format or style. Ignores fundamental constraints. & 
Severe factual inaccuracies or fabricated information. Major logical inconsistencies throughout. Content lacks coherence and structure. Inappropriate or misleading information. & 
Fundamentally unusable response. Multiple critical failures across dimensions. Requires complete reconstruction. Fails to provide meaningful value. \\
\hline
\textbf{2} & 
Partial understanding of user intent. Misses critical elements in requirements. Shows significant gaps in instruction comprehension. Inconsistent adherence to specified constraints. & 
Notable factual errors affecting comprehension. Logical gaps and contradictions present. Limited depth or superficial treatment. Significant portions require correction. & 
Limited utility with significant issues. Substantial revision needed (70\%+ modification). Core problems in execution or understanding. Minimal practical value to user. \\
\hline
\textbf{3} & 
Generally follows instructions with minor deviations. Captures main user intent accurately. Minor non-compliance with secondary requirements. Meets most specified criteria adequately. & 
Generally accurate information with minor flaws. Adequate depth and completeness. Coherent structure and flow. Some areas could benefit from enhancement. & 
Serviceable response meeting basic expectations. Moderate revisions needed (up to 30\% modification). Adequate but unremarkable performance. Provides reasonable value with some limitations. \\
\hline
\textbf{4} & 
Excellent instruction adherence. Addresses all major requirements comprehensively. Demonstrates clear understanding of user needs. \textbf{Maximum score for this dimension.} & 
High-quality, accurate, and comprehensive content. Strong logical consistency. Good depth and relevant details. Well-structured and engaging presentation. & 
High-quality response with notable strengths. Minor adjustments needed (up to 10\% modification). Exceeds basic requirements in multiple areas. Strong practical value and usability. \\
\hline
\textbf{5} & 
\textbf{N/A - Instruction Following capped at 4 points} & 
Exceptional content quality serving as exemplary reference. Expert-level accuracy and insights. Rich, nuanced, and thought-provoking. Demonstrates creativity and originality. & 
Outstanding response serving as benchmark. Minimal or no modification required. Exceptional across all evaluation criteria. Demonstrates innovation, expertise, and excellence. \\
\hline
\end{tabular}
\end{table*}
\textbf{Instruction Following (1-4 scale)} measures how accurately the response follows the given instructions and meets specified requirements. This dimension focuses on:
\begin{itemize}
\item Understanding of user intent and task requirements
\item Compliance with format specifications (word count, structure, style)
\item Adherence to content constraints and guidelines
\item Completion of all requested elements
\end{itemize}

\textbf{Content Quality (1-5 scale)} evaluates the intrinsic quality of the generated content itself. This dimension assesses:
\begin{itemize}
\item Factual accuracy and information reliability
\item Logical flow and coherence of ideas
\item Depth and thoroughness of content coverage
\item Appropriateness and relevance to the topic
\end{itemize}

\textbf{Overall Performance (1-5 scale)} provides a holistic assessment of the response's practical value and user satisfaction. This dimension considers:
\begin{itemize}
\item Practical utility for the intended purpose
\item Amount of editing or revision needed
\item Overall effectiveness in meeting user expectations
\item Integration of instruction following and content quality
\end{itemize}
\begin{table*}[h]
\centering
\caption{WriteEval Dataset Genre and count}
\label{tab:writeeval_genres}
\begin{tabular}{lc|lc}
\toprule
\textbf{Genre} & \textbf{Count} & \textbf{Genre} & \textbf{Count} \\
\midrule
Project Planning & 40 & Copywriting & 37 \\
Official Document Writing & 37 & Composition & 36 \\
Summary Report & 39 & Business Writing & 35 \\
Business Writing & 35 & Social Talk & 34 \\
Plan & 34 & Script & 33 \\
Brainstorming & 31 & Naming & 33 \\
Poetry/Classical Chinese & 31 & Evaluation & 33 \\
Letter & 32 & Article & 32 \\
Teaching Writing & 32 & Titlext & 32 \\
Contract/Agreement & 29 & Report & 28 \\
Folk Text & 27 & Fiction & 27 \\
Technical Document & 26 & Application & 26 \\
Story & 24 & Lecture & 21 \\
Paper & 20 & Legal Document & 20 \\
Lyrics & 13 & Other Genres & 11 \\
\bottomrule
\end{tabular}
\end{table*}
The key distinction between these dimensions is their focus: Instruction Following emphasizes compliance and adherence, Content Quality focuses on the substance and reliability of information, while Overall Performance captures the integrated user experience. A response may score differently across dimensions—for example, perfectly following instructions (high Instruction Following) while containing shallow content (lower Content Quality).

Annotators were trained to evaluate each dimension independently while considering the specific requirements of creative writing tasks.

\subsection{WriteEval Dataset Information}

WriteEval is a comprehensive Chinese creative writing evaluation dataset containing 890 samples collected from real-world scenarios. The dataset covers diverse writing genres and tasks, Each sample includes a writing prompt with specific requirements and reference answers selected by human experts from multiple competitive models.

The dataset construction process involved collecting seed data from authentic writing platforms and augmenting it using self-instruct methodology to maintain realistic writing scenarios. To ensure balanced representation, we employed DeepSeek-V3 to classify samples by genre and adjusted the distribution to match real-world writing task frequencies. The final dataset reflects the actual distribution of creative writing demands encountered in practice.

WriteEval uniformly covers 30 primary writing genres with a total of 377 secondary categories. The dataset includes Chinese-specific genres such as folk texts, classical Chinese, and composition writing alongside universal categories. Table~\ref{tab:writeeval_genres} shows the distribution of samples across major genre categories.Table~\ref{tab:writeeval_sample} show some prompts of  WriteEval dataset.
\begin{figure}
    \centering
    \includegraphics[width=0.7\linewidth]{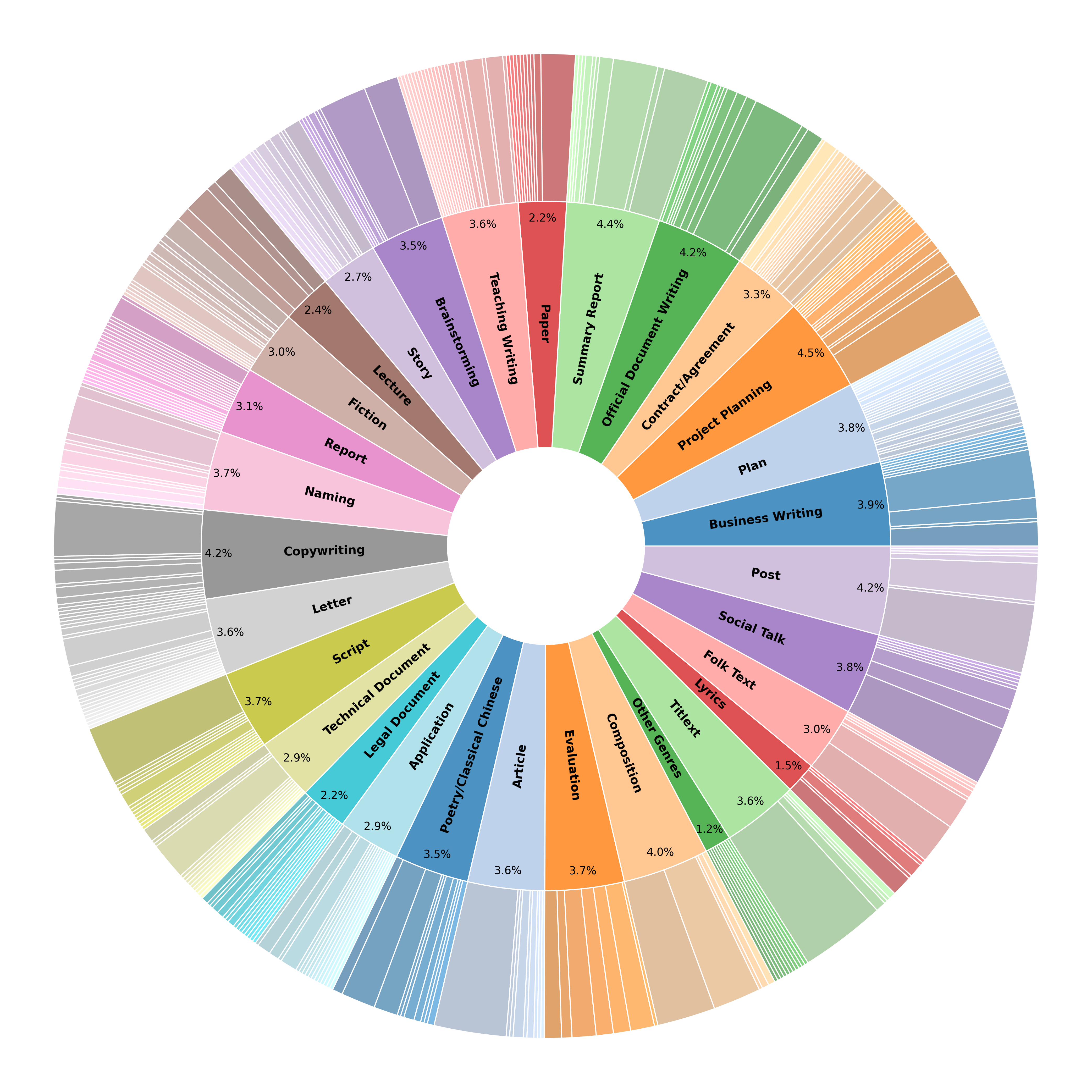}
    \caption{distribution of samples across major genre categories}
    \label{fig:genre}
\end{figure}

\begin{table*}[h]
\centering
\caption{WriteEval Dataset Sample Examples}
\label{tab:writeeval_sample}
\begin{tabular}{p{3cm}p{3cm}p{8cm}}
\toprule
\textbf{Primary Genre} & \textbf{Secondary Genre} & \textbf{Prompt} \\
\midrule
Copywriting & Advertisement Slogan & Design an advertising scenario and a classic slogan for Apple iPhone. No less than 150 words. \\
\midrule
Fiction & Short Story & Write a short story using the following three elements: Tank, Toddler, Fishing Rod \\
\midrule
Business Writing & Business Email & Write a business email introducing the advantages of our bedding sets \\
\midrule
Poetry/Classical Chinese & Modern Poetry & Title: Lotus Root. Reference poem: "New powder by bamboo window / Green grows in lotus pond / Should be in the depths of clouds". Following the style of the above poem, write a three-line poem about "Lotus Root". The word "lotus root" should not appear in the poem. Write it more abundantly. \\
\midrule
Evaluation & Character Evaluation & 12 colleagues have been promoted through job reclassification. Please provide a positive evaluation for them. \\
\bottomrule
\end{tabular}
\end{table*}

\subsection{Case Study}

We present case studies to show how different reward strategies work in practice. These examples help us understand why RLMR performs better than existing methods in real writing tasks.

\subsubsection{Medical Thank-You Letter Reply}

We examine a task where a doctor needs to reply to a patient's thank-you letter. The task has specific requirements: (1) include salutation, greeting, body, closing wishes, signature, and date; (2) start the body with "Thank you very much for your letter, I feel very honored"; (3) end the body with "Thank you again for your recognition and encouragement of my work. Wish you good health and a happy life!".

\begin{figure*}
    \centering
    \includegraphics[width=1\linewidth]{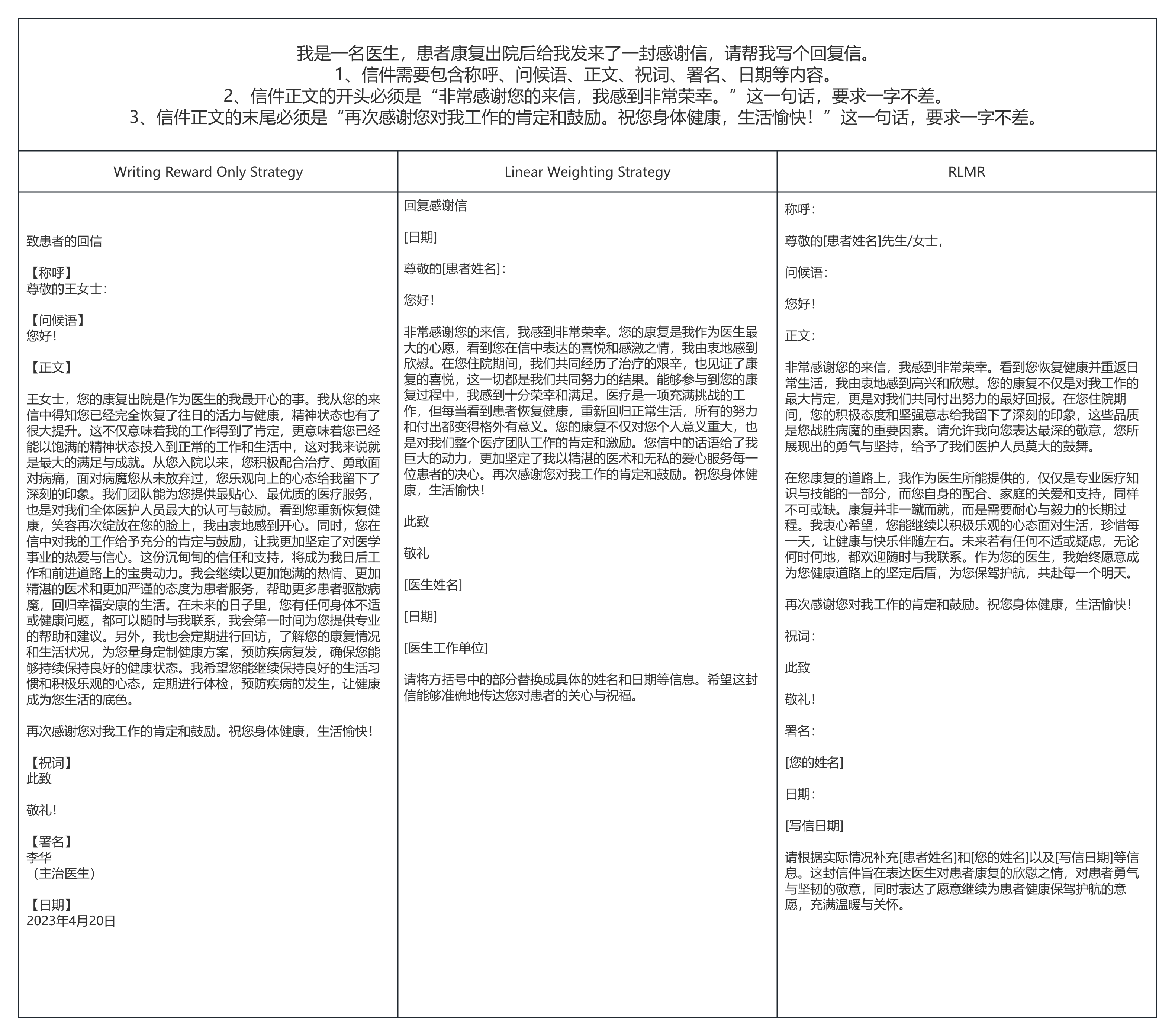}
    \caption{Comparison of three reward strategies for medical thank-you letter reply. RLMR achieves the best balance between content quality and constraint compliance.}
    \label{fig:case_study_medical}
\end{figure*}

As shown in Figure~\ref{fig:case_study_medical}, the three methods produce different results:

\textbf{Writing Reward Only Strategy (3 points):} This approach creates rich and emotionally engaging content with professional warmth. However, it fails to include the required opening phrase "Thank you very much for your letter, I feel very honored", using a generic greeting instead. While the content quality is high, the constraint violation significantly reduces its practical usability.

\textbf{Linear Weighting Strategy (2 points):} This method correctly includes both required opening and closing phrases, showing better instruction following. However, the content between these constraints is overly formulaic and lacks depth. The response reads like a template with minimal personalization, making it feel mechanical despite meeting structural requirements.

\textbf{RLMR (4 points):} Our method achieves good balance between content richness, usability, and formatting requirements. It successfully incorporates both required phrases while maintaining natural, engaging content throughout. The response demonstrates professional expertise and emotional warmth, with smooth content flow that doesn't feel constrained by the requirements.

\subsubsection{CEO Message for Operations Handbook}

We analyze another task where a CEO needs to write a concise message for an operations handbook. The task requires focusing on motivating employees, guiding business learning, overcoming difficulties, and promoting enterprise development, with emphasis on being concise.

\begin{figure*}
    \centering
    \includegraphics[width=1\linewidth]{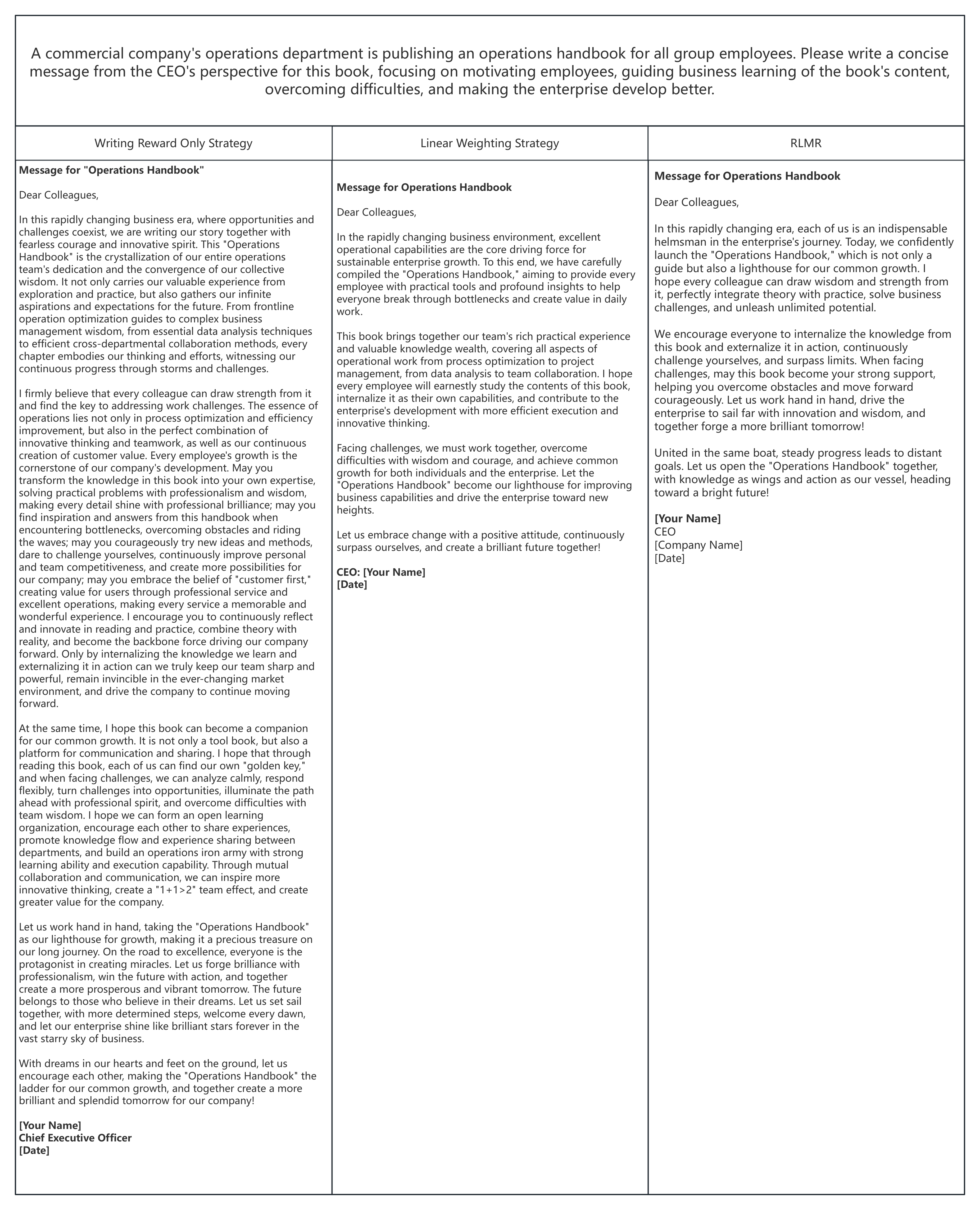}
    \caption{Comparison of three reward strategies for CEO message writing. RLMR delivers concise yet comprehensive content that accurately captures core intentions.}
    \label{fig:case_study_ceo}
\end{figure*}

Figure~\ref{fig:case_study_ceo} presents the outputs from the three approaches:

\textbf{Writing Reward Only Strategy (2 points):} This method produces extremely lengthy content that violates the "concise" requirement. While the content is rich and emotionally engaging, it far exceeds the specified length constraints, making it impractical for actual use. The verbose style undermines the effectiveness of the message.

\textbf{Linear Weighting Strategy (3 points):} This approach achieves better length control but shows content focus deviation. The motivational elements, which should be central to a CEO message, receive insufficient emphasis. While the overall structure is reasonable, the content balance doesn't fully align with the task requirements.

\textbf{RLMR (4 points):} Our method accurately captures the core intentions while maintaining concise and focused content. It successfully balances all required elements - motivation, learning guidance, difficulty overcoming, and enterprise development - within appropriate length constraints. The message is both inspiring and practical, demonstrating effective content organization and priority management.

These case studies reveal how RLMR achieves better balance between writing quality and constraint compliance. Our method helps models follow instructions while maintaining good content, which demonstrates the effectiveness of our dynamic mixed-reward approach in real-world creative writing scenarios.t we aimed for.

\subsection{Prompts Used in Our Work}
\begin{figure*}[htbp]
    \centering
    \includegraphics[width=1\linewidth]{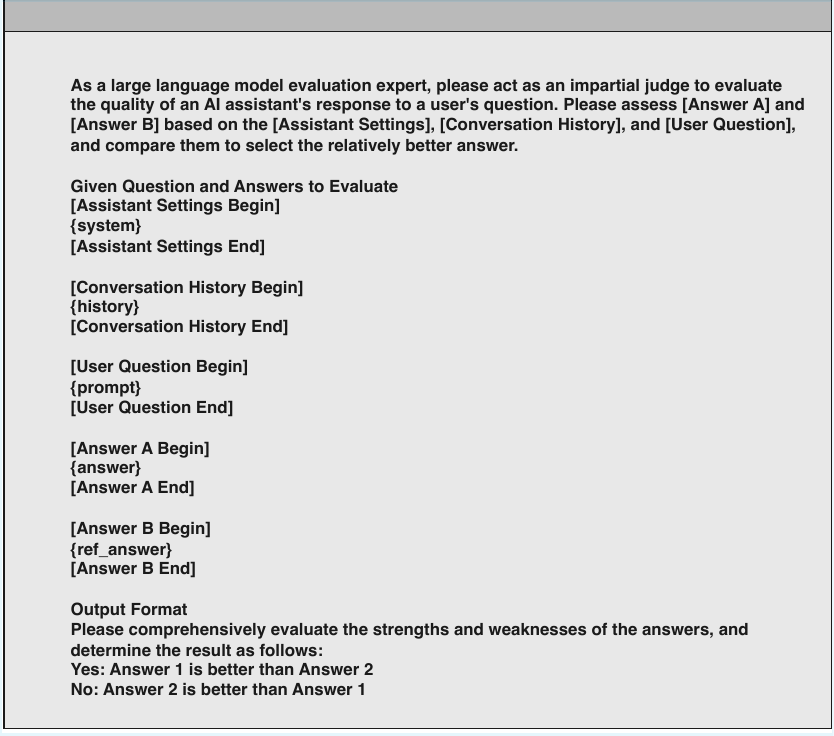}
    \caption{Prompt used for WriteEval automated evaluation}
    \label{fig:grm_prompt}
\end{figure*}

\begin{figure*}[htbp]
    \centering
    \includegraphics[width=0.8\linewidth]{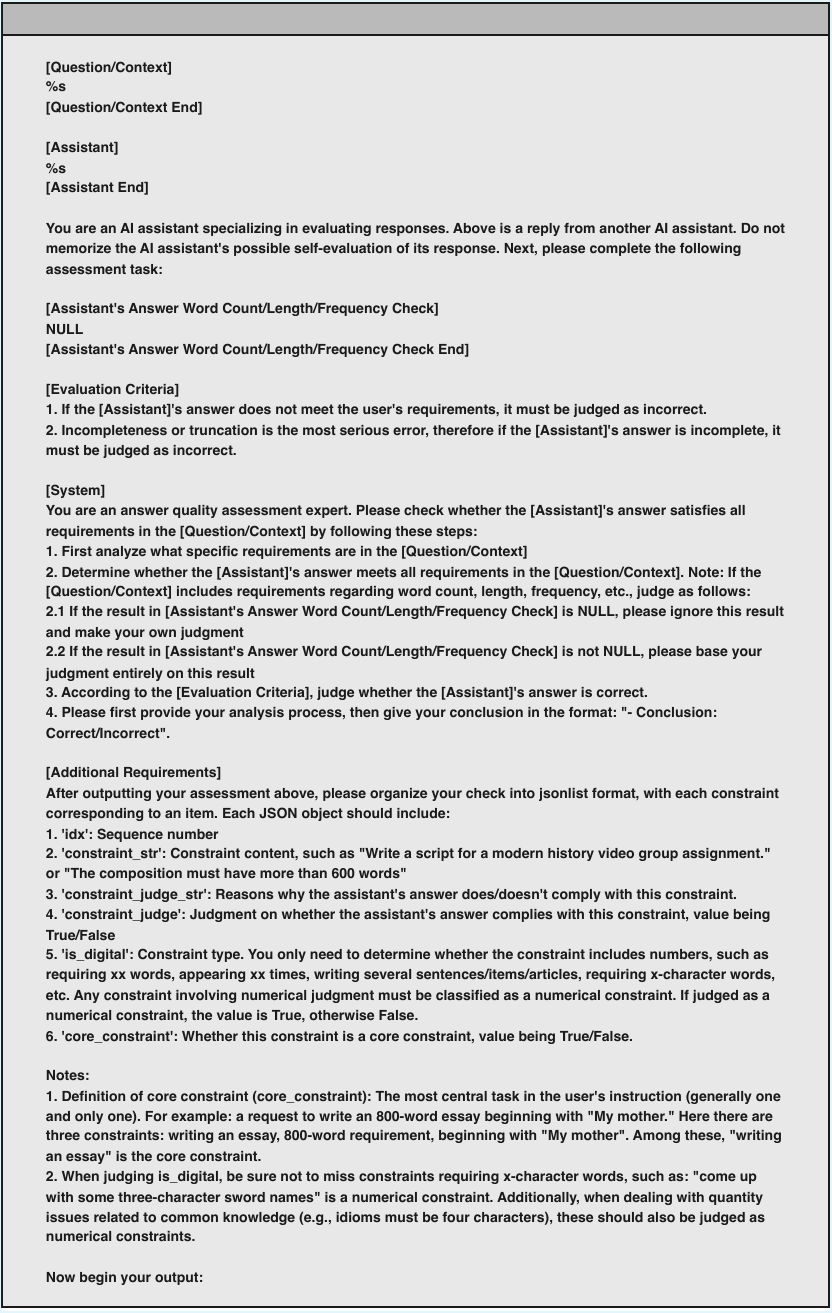}
    \caption{Prompt used by the Constraint Verification Model}
    \label{fig:critic_prompt}
\end{figure*}

The prompts used for WriteEval automated evaluation and Constraint Verification Model are shown in Figures~\ref{fig:grm_prompt} and~\ref{fig:critic_prompt}, respectively.
\subsection{Training Infrastructure and Hyperparameters}

We run on 128 H20 GPUs (64 for GRPO, 64 for services) with 9,743 creative writing queries. We use the VERL framework \cite{sheng2024hybridflow}. Training is 1 epoch (68 steps, 23 hours), learning rate $1 \times 10^{-6}$, batch size 128, 8 samples per query, temperature 1.0, repetition penalty 1.0, max output 14,000 tokens.

\end{document}